\documentclass{article}
\usepackage{spconf,amsmath,graphicx,amssymb}
\usepackage{paralist}
\usepackage{physics}
\usepackage{dsfont}
\usepackage{subfig}
\usepackage{multirow}
\usepackage{mathtools}

\DeclareMathOperator*{\argmax}{argmax}

\title{Alignment Entropy Regularization}
\name{Ehsan Variani, Ke Wu, David Rybach, Cyril Allauzen, Michael Riley}
\address{\{variani, wuke, rybach, allauzen, riley\}@google.com}
\begin{document}
%
\maketitle
\begin{abstract}
Existing training criteria in automatic speech recognition (ASR) permit the model to freely explore
more than one time alignments between the feature and label sequences. In this paper, we
use entropy to measure a model's uncertainty, i.e. how it chooses to distribute
the probability mass over the set of allowed alignments.
Furthermore, we evaluate the effect of entropy regularization in encouraging the model
to distribute the probability mass only on a smaller subset of allowed alignments.
Experiments show that entropy regularization enables a much simpler decoding
method without sacrificing word error rate, and provides better time alignment quality.
\end{abstract}

\section{Introduction}
\label{sec:intro}

Common speech recognition criteria \cite{variani2022global} model $P_\theta(\vb{y}|\vb{x})$, the conditional distribution of
the label sequence $\vb{y}=y_{1:U}$, $y_u\in \Sigma$ given an acoustic feature sequence $\vb{x}=x_{1:T}$, by marginalizing
$P_\theta(\vb{y}, \pi | \vb{x})$ over the alignment $\pi$,
\begin{eqnarray}
P_\theta(\vb{y} | \vb{x}) = \sum_{\pi \in \vb{\Pi}} P_\theta(\vb{y}, \pi | \vb{x})
\label{eq:marginalization}
\end{eqnarray}
where $\vb{\Pi} = \Pi(i_{\theta,\vb{x}}, \vb{y}, F_{\theta,\vb{x}})$ is the set of all the alignment paths corresponding to the label sequence
$\vb{y}$.
$i_{\theta,\vb{x}}$ and $F_{\theta,\vb{x}}$ are respectively the initial and final state of the recognition lattice
$A_{\theta,\vb{x}}$ parameterized by $\theta$.
The choice of the alignment lattice component \cite{variani2022global} determines what alignments constitute $\vb{\Pi}$ and the number of alignments $|\vb{\Pi}|$.
For example, let $T$ be the number of feature frames and $U$ be the number of output labels,
\begin{compactitem}
    \item The listen, attend and spell (LAS) model \cite{chan2015listen} uses the label dependent alignment lattice.
        Each label can be seen as aligned to the entire feature sequence, thus there is single alignment path for any $\vb{y}$ in the recognition lattice.
    \item The connnectionist temporal classification (CTC) \cite{graves2006connectionist} employs the frame dependent alignment lattice.
        Each frame is aligned to at most one label, thus there are $\binom{T}{U}$ alignment paths for any $\vb{y}$ of length $U$.
    \item The recurrent neural network transducer (RNN-T) \cite{graves2012sequence}, and hybrid autoregressive transducer (HAT) \cite{variani2020hybrid} use the label and frame alignment lattice.
    Multiple consecutive labels can align to a single frame, thus there are $\binom{T+U}{U}$ alignment paths for any $\vb{y}$ of length $U$.
\end{compactitem}
In the latter two cases, a recognition lattice admits a large number of alignment paths for the ground truth $\vb{y}$, whereas only a small subset of them are accurate alignments between speech and text.
Because standard training procedures maximize Eq.~\ref{eq:marginalization} without any priors on alignments, a model has a high degree of freedom in distributing $P_\theta(\vb{y}, \pi | \vb{x})$ among different alignments while $P_\theta(\vb{y} | \vb{x})$ remains unchanged.
On one extreme, the model may concentrate the probability mass only on a single alignment (minimum uncertainty).
On the other extreme, it may distribute the mass uniformly over all allowed alignments (maximum uncertainty).
This model uncertainty results in a need for more complex decoding algorithms, and prevents the model from learning accurate alignment between the feature sequence and the label sequence. 

Maximum a posteriori (MAP) inference is typically applied with Eq.~\ref{eq:marginalization}. We first find $\hat{\vb{y}}$ that maximizes $P_\theta(\vb{y}|\vb{x})$, and then find the most likely alignment among all the alignments corresponding to $\hat{\vb{y}}$, in other words,
\begin{align}
\hat{\vb{y}} &= \argmax_{\vb{y}} \,\, \sum_{\pi \in \Pi(i_{\theta,\vb{x}}, \vb{y}, F_{\theta,\vb{x}})} P_\theta(\vb{y}, \pi| \vb{x}) \nonumber \\
\hat{\pi} &= \argmax_{\pi \in \Pi(i_{\theta,\vb{x}}, \hat{\vb{y}}, F_{\theta,\vb{x}})} P_\theta(\vb{\hat{y}}, \pi| \vb{x})
\label{eq:inference}
\end{align}
Since finding the exact $\hat{\vb{y}}$ is usually intractable, a beam search that sums a subset of the identically-label paths is used in practice to find an approximate solution in practice.
We refer to this inference rule as \textit{sum-search} through the paper.
By contrast, a model with less uncertainty concentrates the probability mass only on a few alignment paths or even a single one, making the much simpler maximum alignment probability inference
a good approximation of MAP inference,
\begin{eqnarray}
\hat{\vb{y}}, \hat{\pi} = \argmax_{(\vb{y}, \pi)} \,\, P_\theta(\vb{y}, \pi | \vb{x})
\label{eq:viterbi}
\end{eqnarray}
This single step inference rule is denoted as \textit{max-search}.
For models with a finite label context size $c$ and a small vocabulary $\Sigma$, the most probable alignment path can be efficiently found in $(O(T \cdot |\Sigma|^{c+1})$ time \cite{variani2022global}.
Furthermore, computation on accelerator hardware can benefit from the fixed search
space structure in contrast to the data-dependent structure of beam search \cite{variani2022global}.

The lack of alignment priors in Eq.~\ref{eq:marginalization} means $\hat{\pi}$ from MAP inference may not be an accurate alignment between speech and text.
This is particularly undesirable for tasks relying on high alignment accuracy derived from ASR such as many practical text-to-speech (TTS) models.

In this paper, we propose to calculate alignment entropy
to measure model uncertainty on choice of alignment. 
We also propose to use entropy regularization for
encouraging the model to concentrate more probability mass on fewer alignment paths.
Experiments with both streaming and non-streaming models on the Librispeech corpus \cite{panayotov2015librispeech} show that with entropy regularization, the simpler max-search inference rule of Eq.~\ref{eq:viterbi} can be used without sacrificing word error rate, yet at the same time improve alignment accuracy.

\section{Alignment Entropy}
\label{sec:entropy}

Given the feature sequence $\vb{x}$ and ground truth label sequence $\vb{y}$, we propose to measure model uncertainty through the entropy of distribution $P_\theta(\pi | \vb{x}, \vb{y})$,
\begin{align*}
H_{\theta, \vb{\Pi}}  = & -\sum_{\pi \in \vb{\Pi}} P_\theta(\pi | \vb{x}, \vb{y}) \log(P_\theta(\pi | \vb{x}, \vb{y})) \\
    = & -\frac{1}{P_\theta(\vb{y} | \vb{x})} \sum_{\pi \in \vb{\Pi}} P_\theta(\vb{y}, \pi | \vb{x}) \log(P_\theta(\vb{y}, \pi | \vb{x})) \\
    & + \log(P_\theta(\vb{y} | \vb{x}))
\end{align*}
where $\vb{\Pi}= \Pi(i_{\theta,\vb{x}}, \vb{y}, F_{\theta,\vb{x}})$.
Here we present an iterative derivation of the entropy value for frame dependent alignment lattice.
Similarly, the entropy value can be derived for other type of alignment lattices
which is not presented here due to the space limitation.

Let 
$H_{\theta, \vb{\Pi}_{t,q,u}}$
be the partial entropy defined on all the prefix alignment paths in $\vb{\Pi}_{t,q,u} \triangleq \Pi(i_{\theta,\vb{x}}, y_{1:u}, (t, q))$,
where $(t, q) \in Q_{\theta,\vb{x}}$ is a state in the recognition lattice state space
$Q_{\theta,\vb{x}}$. Note that $H_{\theta, \vb{\Pi}} = H_{\theta, \vb{\Pi}_{T,F,U}}$ by definition.
\begin{eqnarray}
H_{\theta, \vb{\Pi}_{t,q,u}} &=& -\sum_{\mathclap{\substack{\pi \in \Pi(i_{\theta,\vb{x}}, y_{1:u}, (t, q))}}} p_{\pi} \log (p_{\pi}) \nonumber \\
&=& -\sum_{\pi} \frac{\omega[\pi]}{\alpha_{\vb{\Pi}_{t,q,u}}} \log (\frac{\omega[\pi]}{\alpha_{\vb{\Pi}_{t,q,u}}}) \nonumber \\
&=& \frac{-1}{\alpha_{\vb{\Pi}_{t,q,u}}} \underbrace{\left[\sum_{\pi} \omega[\pi] \log (\omega[\pi]) \right]}_{= A_{\vb{\Pi}_{t,q,u}}} \nonumber \\
&+& \log (\alpha_{\vb{\Pi}_{t,q,u}})
\label{eq:partial_entropy}
\end{eqnarray}
where $p_{\pi} \triangleq \omega[\pi] / \alpha_{\vb{\Pi}_{t,q,u}}$ is the probability of path $\pi$,
with path weight $\omega[\pi]$ and $\alpha_{\vb{\Pi}_{t,q,u}} = \sum_{\pi} \omega[\pi]$.
\begin{eqnarray}
A_{\vb{\Pi}_{t,q,u}} &=& \sum_{\pi \in \Pi_1, e_1} (\omega[\pi] \omega[e_1]) \log (\omega[\pi] \omega[e_1]) \nonumber \\
&+& \sum_{\pi \in \Pi_2, e_2} (\omega[\pi] \omega[e_2]) \log (\omega[\pi] \omega[e_2]) \nonumber
\end{eqnarray}
where
\begin{eqnarray}
\Pi_1 &=& \Pi(i_{\theta,\vb{x}}, y_{1:u-1}, (t-1, p)) \nonumber \\
e_1 &=& ((t-1, p), y_u, \omega_{\theta,\vb{x}}(t-1, p, y_u), (t, q)) \nonumber \\
\Pi_2 &=& \Pi(i_{\theta,\vb{x}}, y_{1:u}, (t-1, q)) \nonumber \\
e_2 &=& ((t-1, q), \epsilon, \omega_{\theta,\vb{x}}(t-1, p, \epsilon), (t, q)) \nonumber \\
\nonumber
\end{eqnarray}
and $(t-1, p)$ is any prefix state of $(t, q)$ in the recognition lattice $A_{\theta,\vb{x}}(\vb{y})$.
The above equation can be further simplified as:
\begin{eqnarray}
A_{\vb{\Pi}_{t,q,u}} &=& \omega[e1] \biggl[ A_{\vb{\Pi}_{t-1,p,u-1}} + \log (\omega[e_1]) \biggr] \nonumber \\
&+& \omega[e2] \biggl[ A_{\vb{\Pi}_{t-1,q,u}} + \log (\omega[e_2]) \biggr]
\end{eqnarray}
Thus $A_{\vb{\Pi}_{t,q,u}}$ can be calculated iteratively and so $H_{\vb{\Pi}_{t,q,u}}$ in Eq.~\ref{eq:partial_entropy}.

\subsection{Implementation}
The main implementation consideration is the need to deal with the underflow problem which
happens when there is an unlikely path with too small probability. To avoid
underflow, we implemented the entropy derivation in log domain which respectively
requires representing log probabilities in log-of-log domain.
Empirically we found that
calculating entropy in the log domain is sufficient for addressing numerically instability of the algorithm due to
the underflow.

\subsection{Maximum Entropy}
The maximum entropy is achieved when model uniformly distributes the conditional probability
mass over alignments i.e, $P_{\theta}(\vb{y}, \pi | \vb{x}) = P_{\theta}(\vb{y} | \vb{x}) / N$ where $N$ is
the total number of alignments. Thus 
\[
P_\theta(\pi | \vb{x}, \vb{y}) = \frac{P_{\theta}(\vb{y}, \pi | \vb{x})}{P_{\theta}(\vb{y} | \vb{x})} = 1/N
\]
and the maximum entropy for frame dependent alignment lattice is then equal to:
\begin{eqnarray}
H_{max} &=& -\sum_{\pi \in \vb{\Pi}} 1/N \log(1/N) = \log({T \choose U})
\end{eqnarray}

\subsection{Regularization}

We can regularize model with the alignment entropy by optimizing the following loss:
\begin{eqnarray}
\mathcal{L}(\theta) = E_{P(\vb{x},\vb{y})}[-\log P_{\theta}(\vb{y}|\vb{x}) + \lambda H_\theta(\pi) ]
\label{eq:alignment_entropy_regularization}
\end{eqnarray}
with samples from the true data distribution $P(\vb{x}, \vb{y})$. The regularization parameter
can be swept through a range of values.

\section{Experiments}
\label{sec:experiments}

The experiments are designed to evaluate streaming and non-streaming models with
respect to the alignment entropy, alignment timing quality as well as comparison of the sum-search
and the max-search inference rules.
These are all investigated for models trained with or without alignment entropy regularization.

\textbf{Data \quad}
The full $960$-hour Librispeech corpus is used \cite{panayotov2015librispeech} for experiments.
The input features are $80$-dim. log Mel extracted from a $25$ ms window of the speech signal with a $10$ ms shift.
The SpecAugment library with baseline recipe parameters were used \cite{gulati2020conformer}.
The ground truth transcription is used without any processing and tokenized by the $28$ graphemes
that appear in the training data.

\textbf{Architecture \quad}
The architecture used for all experiments is exactly matching the shared-rnn architecture
in \cite{variani2022global}. It consists of an acoustic encoder
and a label encoder.
The acoustic encoder is a stack of $12$-layer Conformer encoders \cite{gulati2020conformer} with
model dimension $512$, followed by a linear layer with output dimension $640$.
The Conformer parameters are set such that the only difference between streaming
and non-streaming models is the right context:
at each time frame $t$, the streaming models only access the left context (feature frames from $1$ to $t$),
while the non-streaming models can see the entire
acoustic feature sequence. More details about the architecture can be found in \cite{variani2022global}.
The label encoder is a single layer LSTM with $640$ cells.
All the experiments use context size $c=2$, which means the LSTM state
is reset every two labels.
All experiments are locally normalized with HAT factorization \cite{variani2020hybrid}.

\textbf{Training \quad}
All models are trained on $8\times8$ TPUs with a batch size $2048$.
The training examples with more than $1961$ feature frames or more than
$384$ labels are filtered out. We used Adam optimizer
\cite{kingma2014adam} ($\beta_1=0.9$, $\beta_2=0.98$,
and $\epsilon=10^{-9}$) with the Transformer learning rate schedule 
\cite{vaswani2017attention} (10k warm-up steps and peak learning 
rate $0.05/\sqrt{512}$). We applied the same regularization techniques
and the training hyper parameters used in the baseline recipe of \cite{gulati2020conformer}.

Four model configurations were trained which are different whether they are streaming
or not and also if they are trained with or without alignment entropy regularization.
All models trained with the criterion in Eq.~\ref{eq:alignment_entropy_regularization}. For
baseline models $\lambda$ is set to zero and models trained with alignment regularization
use $\lambda=0.01$ which was found empirically by sweeping over range of values.

\textbf{Evaluation \quad}
For recognition quality we report word error rate (WER) on the standard Librispeech test
sets: test\_clean and test\_other.
The checkpoint with the lowest WER obtained from the geometric mean between
the dev\_clean and dev\_other sets was used for evaluation on the
test\_clean and test\_other sets.

For time alignment quality, we compared the word alignment between top alignment path
$\hat{\pi}$ and the reference word alignment obtained from Montreal forced aligner tool
\cite{mcauliffe2017montreal}. For comparison we report alignment accuracy $\text{ACC}(\tau)$ for
different threshold value $\tau$ which is calculated as follow:
For hypothesis $h$ and reference $r$:
\begin{enumerate}
    \item For each word $w \in h$, define $h_s(w)$ to be the time stamp that the first letter of $w$ is recognized.
    Similarly define $h_e(w)$ to be the time stamp that last letter of $w$ is recognized. The corresponding values
    for reference sequence are denoted as $r_s(w)$ and $r_e(w)$, respectively.
    \item The alignment accuracy for threshold $\tau$ is then defined as:
    \[
    \text{ACC}(\tau) = \frac{\sum_w \mathds{1}(r_s(w) - \tau 
    \leq h_s(w) \land  h_e(w) 
    \leq r_e(w) + \tau)}{N_w}
    \]
    where summation is over all the words in the test set and $N_w$ is the total number of words. 
\end{enumerate}

\begin{figure}[t]
  \centering
  \includegraphics[width=\linewidth]{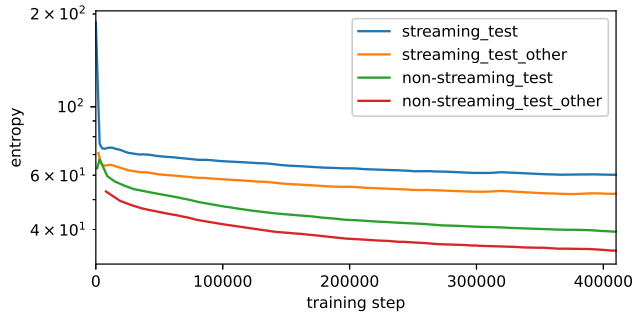}
  \caption{Alignment entropy over course of training.}
  \label{fig:baseline_entropy}
\end{figure}

\begin{figure}[h]
  \centering
  \includegraphics[width=\linewidth]{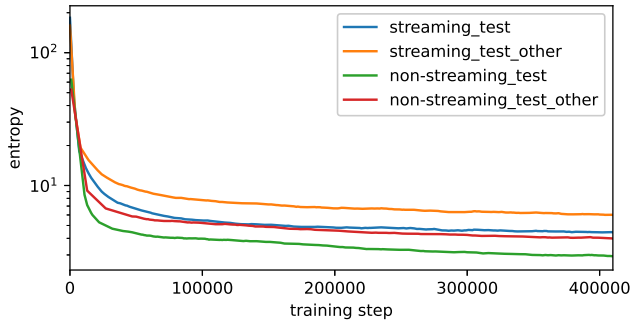}
  \caption{Alignment entropy over course of training. All models are trained with alignment entropy regularization.}
  \label{fig:regularized_entropy}
\end{figure}

\textbf{Alignment path entropy \quad}
Figure~\ref{fig:baseline_entropy} shows how the alignment entropy changes over training time for
both streaming and non-streaming models. 
For comparison, the maximum entropy on test\_clean and test\_other are $218.3$ and $189.8$, respectively.
At the beginning of training, the entropy value is close to the maximum entropy
which express model's uncertainty on choice of alignment and it uniformly distributes its probability mass across all alignments. As training proceeds,
the entropy goes down which imply that model start concentrating on subset of paths. 
The non-streaming model entropy is much smaller than its streaming counterpart.
Figure~\ref{fig:regularized_entropy} presents same training curves when alignment entropy regularization
is applied. Similar to baseline models, model start with relatively large entropy
value and eventually converge to much smaller value. The final entropy value is significantly
smaller for all models compare to Figure~\ref{fig:baseline_entropy}.
Furthermore, the difference between streaming and non-streaming models is also shrinking when
model is trained with regularization.

\begin{figure}[t]
    \centering
    \subfloat[\centering Without regularization]{{\includegraphics[height=3cm,width=3.85cm]{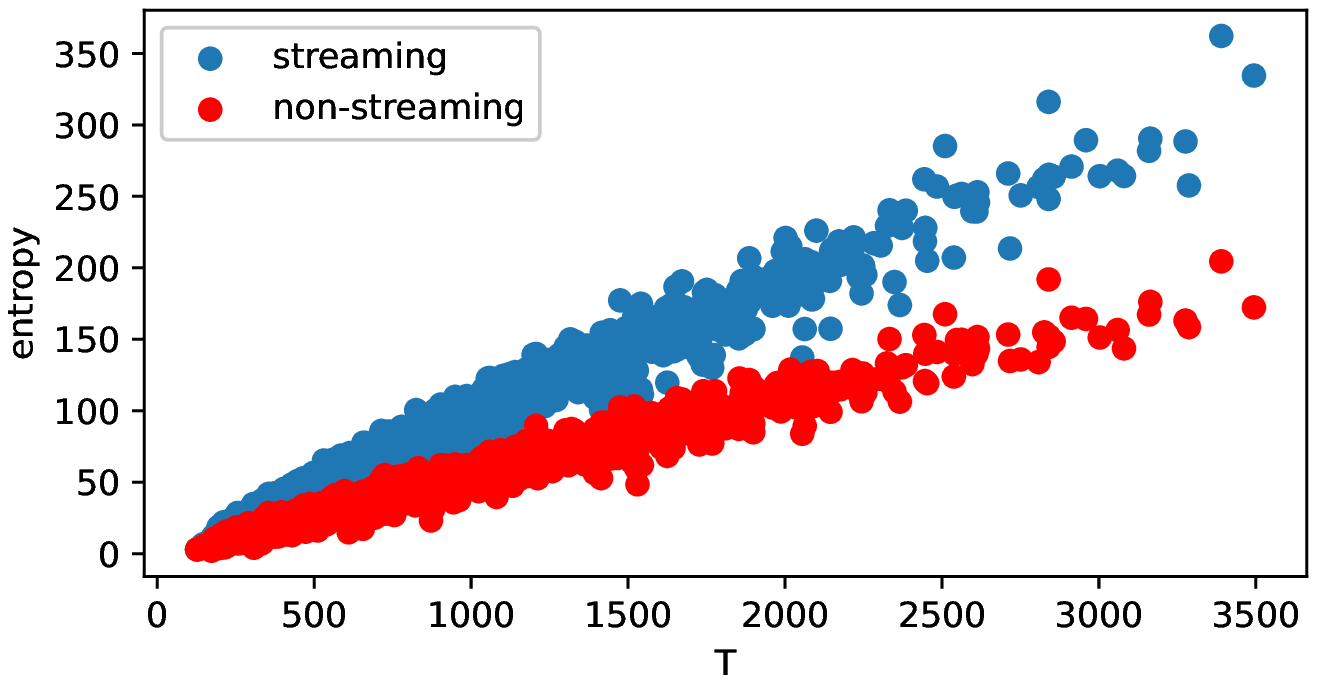} }}%
    \qquad
    \subfloat[\centering With regularization]{{\includegraphics[height=3cm,width=3.85cm]{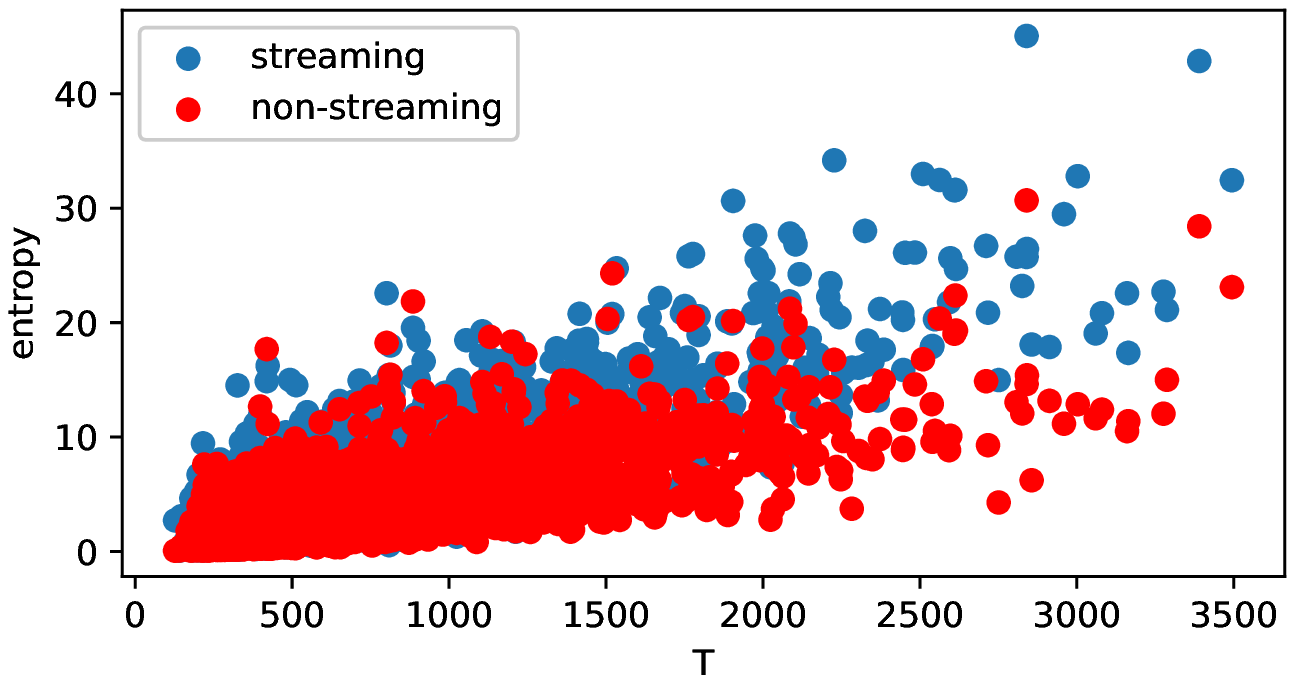} }}%
    \caption{Scatter plot of entropy values (y-axis) per number of frames (x-axis) for streaming and non-streaming models
    trained without or with alignment entropy regularization.}%
    \label{fig:scatter}%
\end{figure}

The entropy value versus number of time frames is presented in Figure~\ref{fig:scatter}.
For baseline models, the entropy value linearly scales by the number of frames. The scale is
relatively smaller for non-streaming models compare to the streaming ones. Applying alignment
entropy regularization lead to much smaller entropy value for both short and long sequences
whether model is streaming or not, Figure~\ref{fig:scatter}-b.

\textbf{Choice of inference rule \quad}
Table~\ref{tab:wer} compare sum-search and max-search inference rules across different models. For baseline models, the sum-search rule achieve better WER in general. The difference is about $0.1 \%$ WER in all cases except for the streaming model on test\_other set. There max-search performs $11.1 \%$ while the sum-search
achieves $10.0 \%$. Note that when alignment entropy regularization is applied, the max-search rule achieves similar performance as sum-search rule across the board. This is encouraging since as mentioned before, the max-search rule is an exact search for wide range of models, and can be efficiently implemented on accelerators as explained in \cite{variani2022global}.

\begin{table}[h]
\centering
    \begin{tabular}{|c|c|c|c|c|c|}
    \hline
    {test} & streaming &  {entropy} & Entropy & \multicolumn{2}{c|}{inference rule} \\
    \cline{5-6}
    set    &           & reg.       &               &  sum  &  max\\
    \hline
    \multirow{4}{*}{clean} & \multirow{2}{*}{no} & no  & 39.6 & 2.5 & 2.5\\
    \cline{3-6}
                           &                     & yes & 2.8 & 2.5  & 2.6\\
    \cline{2-6}
                           & \multirow{2}{*}{yes} & no  & 60.1 & 4.6  & 4.7\\
    \cline{3-6}
                           &                     & yes & 4.4  & 4.4  & 4.4\\
    \hline
    \multirow{4}{*}{other} & \multirow{2}{*}{no} & no  & 34.7 & 5.8  & 5.8\\
    \cline{3-6}
                           &                     & yes & 3.9 & 5.8  & 5.8\\
    \cline{2-6}
                           & \multirow{2}{*}{yes} & no  & 52.2 & 10.0 & 11.1\\
    \cline{3-6}
                           &                     & yes & 6.0  & 10.0   & 10.0\\
    \hline
    \end{tabular}
    \caption{\label{tab:wer} The convergence entropy and WER for streaming and non-streaming models trained with or without alignment entropy regularization.}
\end{table}

\textbf{Time alignment quality \quad}
We compared the time alignment quality by calculating the alignment accuracy $ACC(\tau)$
over different threshold values $\tau$.
As shown in Table~\ref{tab:non-streaming-acc}, the non-streaming models trained with regularization, achieve much
better alignment accuracy for smaller threshold values. Specifically, for $\tau=0$, the regularized model gains about $22 \%$ relative alignment accuracy. The gain is smaller as the threshold increases and for threshold $\tau=50$ msec, both baseline and regularized model perform equally well.
\begin{table}[!h]
\centering
    \begin{tabular}{|c|c|c|c|c|c|c|c|}
    \hline
    {test} & {entropy} & \multicolumn{6}{c|}{$\tau$ [msec] } \\
    \cline{3-8}
    set & reg. & 0 & 10 & 20 & 30 & 40 & 50 \\
    \hline
    \multirow{2}{*}{clean} & {no} & 66  & 77 & 86  & 93 & 96  & 98\\
    \cline{2-8}
                     & yes & \bf 81 & \bf 87 & \bf 93 & \bf 97 & \bf 99 & \bf 99\\
    \hline
    \multirow{2}{*}{other} & {no} & 61 & 72 & 83 & 90 & 95 & 97\\
    \cline{2-8}
                     & yes  & \bf 77 & \bf 85 & \bf 91 & \bf 95 & \bf 98 & \bf 99\\
    \hline
    \end{tabular}
    \caption{\label{tab:non-streaming-acc} Alignment accuracy for non-streaming model.}
\end{table}

The alignment accuracy for streaming models is shown in Table~\ref{tab:streaming-acc}. The streaming models
achieve significantly lower alignment accuracy compare to the non-streaming counterpart. Similar to Table~\ref{tab:non-streaming-acc},
models trained with alignment entropy regularization achieve better alignment accuracy compare to the baseline
models, however the gain is relatively smaller compare to the non-streaming models.
\begin{table}[h]
\centering
    \begin{tabular}{|c|c|c|c|c|c|c|c|c|}
    \hline
    {test} & {entropy} & \multicolumn{7}{c|}{$\tau$ [msec] } \\
    \cline{3-9}
    set & reg. & 0 & 100 & 200 & 300 & 400 & 500 & 600 \\
    \hline
    \multirow{2}{*}{clean} & {no} & 0  & 2 & 7  & 17 & 39  & 86 & 100\\
    \cline{2-9}
                     & yes & \bf 1 & \bf 3 & \bf 10 & \bf 22 & \bf 46 & \bf 89 & \bf 100\\
    \hline
    \multirow{2}{*}{other} & {no} & 0 & 1 & 6 & 5 & 34 & 83 & 100\\
    \cline{2-9}
                     & yes  & \bf 1 & \bf 2 & \bf 8 & \bf 18 & \bf 40 & \bf 87 & \bf 100\\
    \hline
    \end{tabular}
    \caption{\label{tab:streaming-acc} Alignment accuracy for streaming model.}
\end{table}


\vspace{-0.3cm}
\section{Conclusion}
\label{sec:conclusion}

We presented a derivation of alignment path entropy as a measure for evaluating model's uncertainty
on choice of alignment. We showed that both streaming and non-streaming models are intended to
distribute their probability mass over relatively large subset of alignment paths which can lead
to some undesired inference complexity and time alignment quality. It was shown that with
alignment entropy regularization, we can use much simpler and exact inference rule without loss
of performance quality. Furthermore, alignment entropy regularization can significantly improve
the time alignment accuracy.

\vfill
\pagebreak

\bibliographystyle{IEEEbib}
\bibliography{strings,refs}

\end{document}